\documentclass[letterpaper, 10 pt, conference]{ieeeconf}  %

\IEEEoverridecommandlockouts                              %

\usepackage{graphicx} %
\usepackage{amsmath} %
\usepackage{amssymb}  %
\usepackage{xcolor}
\usepackage[T1]{fontenc}  %

\usepackage{balance}
\usepackage{siunitx}

\usepackage{color} %
\usepackage{algorithm,algorithmic} %

\title{\LARGE \bf
	Offline Object Extraction from \\ 
	Dynamic Occupancy Grid Map Sequences$^\star$\
}

\author{Daniel Stumper$^{1}$, Fabian Gies$^{1}$, Stefan Hoermann$^{1}$, and Klaus Dietmayer$^{1}$%
\thanks{$^{1}$All authors are with the Institute of Measurement, Control and Microtechnology, Ulm University, Germany.
	{\tt\small firstname.lastname@uni-ulm.de}
	$^\star$"This work has been submitted to the IEEE for possible publication. Copyright may be transferred without notice, after which this version may no longer be accessible"}%
}

\newcommand{\EMAGSlong}{{\textbf e}go {\textbf m}otion {\textbf a}ligned {\textbf g}rid map {\textbf s}equence}
\newcommand{\EMAGS}{EMAGS}

\newcommand{\initpoint}{initialization point}
\newcommand{\initpoints}{initialization points}
\newcommand{\InitPoint}{Initialization Point}

\pdfoutput=1

\begin{document}
\def\cred{\textcolor{red}}
\def\cblue{\textcolor{blue}}
\def\cgreen{\textcolor{green}}

\def\GMchannels{\Omega}
\def\GMwidth{W}
\def\GMheight{H}

\def\GMCellOrientation{\theta}
\def\OBJOrientation{\phi}
\def\Occupied{\mathrm{O}}
\def\MassFree{M_\mathrm{F}}
\def\MassOcc{M_\mathrm{O}}
\def\East{\mathrm{E}}
\def\North{\mathrm{N}}
\def\EorN{*}

\def\static{\mathrm{s}}
\def\dynamic{\mathrm{d}}

\maketitle
\thispagestyle{empty}
\pagestyle{empty}

\begin{abstract}
	A dynamic occupancy grid map (DOGMa) allows a fast, robust, and complete environment representation for automated vehicles.
	Dynamic objects in a DOGMa, however, are commonly represented as independent cells while modeled objects with shape and pose are favorable.
	The evaluation of algorithms for object extraction or the training and validation of learning algorithms rely on labeled ground truth data. 
	Manually annotating objects in a DOGMa to obtain ground truth data is a time consuming and expensive process.
	Additionally the quality of labeled data depend strongly on the variation of filtered input data. The presented work introduces an automatic labeling process, where a full sequence is used to extract the best possible object pose and shape in terms of temporal consistency. 
	A two direction temporal search is executed to trace single objects over a sequence, where the best estimate of its extent and pose is refined in every time step. Furthermore, the presented algorithm only uses statistical constraints of the cell clusters for the object extraction instead of fixed heuristic parameters. Experimental results show a well-performing automatic labeling algorithm with real sensor data even at challenging scenarios. 
\end{abstract}

\section{Introduction}
\label{sec_introduction}
For automated driving or modern driver assistant systems, a detection of the vehicle surrounding is essential. For that reason, more and more sensors are mounted on the vehicle to generate dense and precise measurements of the environment. A well-studied topic to detect and track external dynamic objects in the environment is using temporal filtering algorithms \cite{BarShalom1988}. These \textit{object-model-based} representations use Bayesian filtering techniques and manage to suppress clutter and false alarms, and are able to track multiple objects at once \cite{Mahler2007, Reuter2015}. Despite the impressive success, object tracking in crowded urban shared space scenarios is still an tough challenge. 
Using occupancy grid maps is a complementing alternative to process sensor measurements and represent the complete environment \textit{object-model-free} \cite{Elfes1989}. 
Therefore, the local environment is separated in grid cells, where the state of each cell is an estimation of the probabilities for occupied and free. 
The extension to a dynamic occupancy grid map (DOGMa) \cite{Danescu2011,Nuss2017,Rexin2017} enables the estimation of dynamics in any cell.
A main advantage of grid maps is the simple accumulation of sensor measurements from different sensors at different time steps. 
Each cell is processed independently without any assumptions of object shapes, movements or types. 
Consequently, a detection of the entire environment including all traffic participants is possible. 

Due to the independence of cells, there is no information of the associated object generating these measurements. 
However, for autonomous applications, e.g. behavior planning \cite{Ulbrich2013}, full knowledge of the single object state is favorable. 
To achieve this, a major challenge is to extract objects from the grid map by associating cells to objects and represent them with spatial and dynamic information.

\begin{figure}[t]
	\centering
	\includegraphics[width=0.9\linewidth]{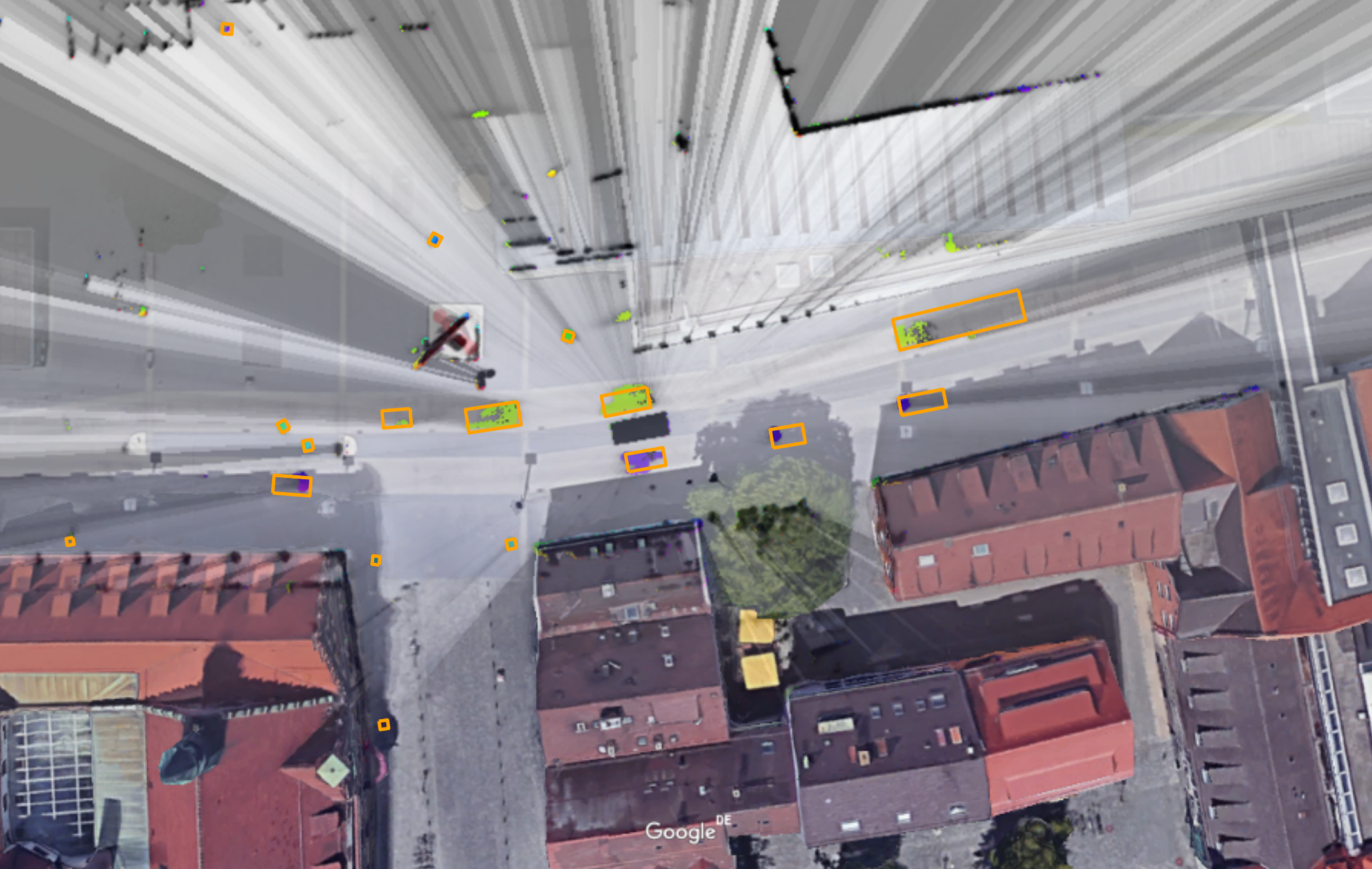}
	\caption{Overlay of the DOGMa and Google Maps in Ulm inner city with extracted objects as orange rectangles after the full sequence is processed.}
	\label{fig_teaser}
\end{figure}

In this work we present an offline approach to extract dynamic objects from a DOGMa. 
Therefore, the presented algorithm uses acausal information from the future and past to generate a ground truth object state to any time. 
Starting from a moment where an object is clearly visible, it can be traced forward and backward in time, while the correct shape, pose and trajectory is refined via best fit on the entire sequence. 
Due to this algorithm, even challenging separations of objects moving next to each other and precise spatial information of occluded or barely visible objects are possible. 
An example of the algorithm's result is shown in Figure \ref{fig_teaser}. Therefore, an overlay of the satellite image from Google Maps and a DOGMA generated with Lidar sensors at an urban area is depicted. Rectangles represent the extracted objects pose and shape, as width and length, after the presented two directional search. It is clearly visible that even objects with only a few number of cells hold the exact estimation of the shape. 
Additionally the algorithm is designed to require almost no heuristic parameters and uses statistical constraints instead. 
The main goal of the algorithm's result is intended to serve as ground truth to evaluate object extraction algorithms, e.g. based on clustering techniques like DBSCAN \cite{Ester1996}, or as labels for learning techniques on grid maps what gains interest in current research \cite{Piewak2017, Hoermann2017}.

The remaining paper is structured as follows: A short insight of related work for objects extraction of grid maps is reviewed in Section \ref{sec_relatec_work}. An implementation of the DOGMa and a prepossessing of the algorithm is described in Section \ref{sec_preprocessing}. The object extraction algorithm with its detailed description is given in  Section \ref{sec_algo_overview} and Section \ref{sec_algo_components}. Resulting extracted objects from the presented algorithm and limitations are shown in Section \ref{sec_results} followed by conclusions given in Section \ref{sec_conclusion}. 

\section{Related Work}
\label{sec_relatec_work}
A common approach to extract objects from the occupancy grid map is based on a combination of multi-object tracking algorithms. In \cite{Bouzouraa2010} a fusion approach is presented where a Kalman filter processes the cell states to improve the object tracking estimate. Therefore, an association between the grid map cells and the current track is necessary, what is realized with a grouping algorithm using a distance criterion. The approach by Jungnickel  \cite{Jungnickel2014} presents an object tracking based only on occupancy grid maps. A particle filter tracks a dynamic cell cluster what represents an arbitrary object shape. For state estimation the DBSCAN algorithm clusters dynamic cells and to sets up new object tracks. 
Schütz et al. \cite{Schuetz2014} perform an extended object tracking \cite{Granstroem2016} using a local grid to estimate the objects shape. Here, the DBSCAN algorithm is also used to cluster the measurements by defining a range parameter $\epsilon$ and minimum number of cells $k_{min}$. Recently published approaches by \cite{Steyer2017,Yuan2017} seem very promising for detecting objects and tracking the pose and shape of objects. However, setting up new objects requires well separable clusters and small uncertainties in the cells. Additionally, heuristic parameter tuning is commonly required and strongly dependent on the density in the scene. Summarized, all online object tracking approaches suffer from engineered feature selections and parameter adjustments.
In \cite{Piewak2017, Hoermann2017} CNNs were trained on DOGMa input to detect and predict objects, while the objects are still represented as single independent cells, rather than clusters or boxes.
Nevertheless, hours of training data, that commonly is labeled manually, is required to use neural networks efficiently. This procedure is expensive and time intensive for a huge amount of data. Due to offline processing, it is possible to automatically label ground truth data by using a two direction temporal search. 
An object detection algorithm, i.e. detecting rotated bounding boxes in a DOGMa, trained with the result presented in this work was published in \cite{Hoermann2018GMObjectDetection}.
\section{Preprocessing}
\label{sec_preprocessing}
The DOGMa is an implementation of \cite{Nuss2017}, where cells~$c$ discretize the local environment as spatial grid at the Universal Transverse Mercator (UTM) coordinates $ (\East,\North) $. The spatial grid provides cells in $\mathbb{R}^{\GMwidth \times \GMheight}$ with width~$\GMwidth$ and height~$\GMheight$ pointing east and north, respectively. A particle filter estimates the static and dynamic state per cell. 
A cell comprises $\GMchannels_c = \left\{ \MassOcc, \MassFree, v_\East, v_\North, \sigma^2_{v_\East}, \sigma^2_{v_\North}, \sigma^2_{v_\East, v_\North} \right\}$ with the Dempster Shafer \cite{Dempster1968} masses for occupancy $\MassOcc \in \left[0,1\right]$ and free space $\MassFree \in \left[0,1\right]$. 
Furthermore, a velocity in east $v_\East$ and north $v_\North$ direction with appropriate \mbox{(co-)variances} $ \left\{\sigma^2_{v_\East}, \sigma^2_{v_\North}, \sigma^2_{v_\East, v_\North} \right\}$ is estimated. 
The grid map cell states $\GMchannels_c(t)$ are given at any time step $t$ of the sequence.
The occupancy probability of a cell is calculated by $P_{\Occupied} = 0.5\cdot\MassOcc + 0.5\cdot(1-\MassFree)$.

\begin{figure}[t]
	\vspace{1.5mm}
	\centering
	\includegraphics[width=0.9\linewidth]{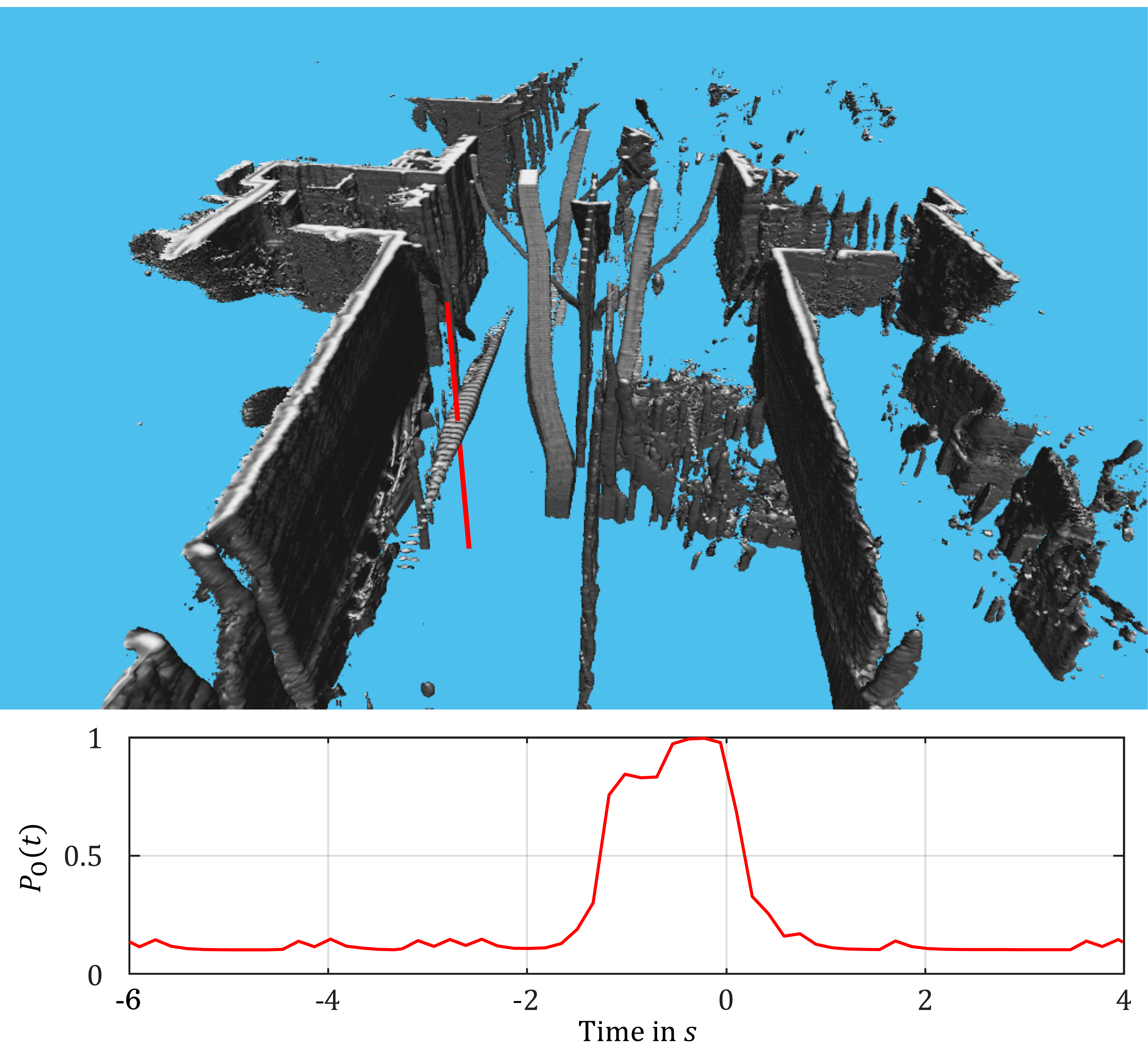}
	\caption{Illustrated \EMAGS{} (top) where each slice is another time step, stacked on top of the previous time step. Static objects, such as walls or pillars, can be seen as vertical objects. Moving objects appear similar to staircases.
	The red vertical line (top) and curve (bottom) illustrate the occupancy of a single cell traversed by an object.}
	\label{fig_emags-1}
\end{figure}

The input data for the algorithm is the \emph{\EMAGSlong{}} (\EMAGS{}) which is a stack of temporal excerpts from a DOGMa sequence.
It is generated by aligning snapshots from the DOGMa according to the ego motion of the perceiving vehicle, to generate a persistent map along the sequence. 
Therefore, even the ego vehicle generates a moving object in the \EMAGS{}, but static objects stay on the same position over time.
Additionally, this implies that every slice in the \EMAGS{} may have other spatial boundaries, depending on the ego motion.
The \EMAGS{} is illustrated in Fig.~\ref{fig_emags-1}, where
static objects can be seen as vertical objects, while moving objects appear similar to a staircase.
In the image, a red line is drawn along the time axis with constant cell coordinates.
The according curve $P_{\Occupied}(t)$ is given in the plot in Fig.~\ref{fig_emags-1}.

\begin{figure}[t]
	\vspace{1.5mm}
	\centering
	\includegraphics[width=0.9\linewidth]{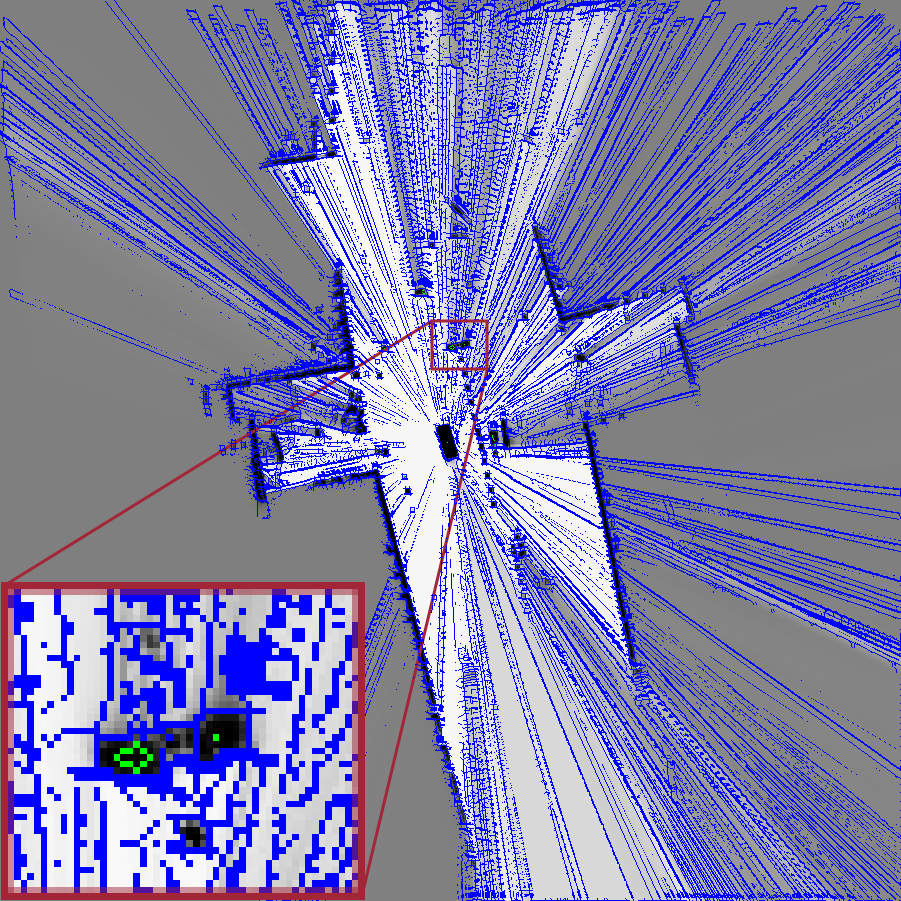}
	\caption{One time step of the preprocessed data is shown, referring to one slice from the \EMAGS{}. The occupancy information is used as background, where black means occupied and white means not occupied. The border mask is shown in blue lines and the calculated \initpoints{} are plotted in green. A zoomed excerpt is integrated in the bottom left, including two possible objects. One is marked with a single \initpoint{}, the other with multiple. A third occupied area is seen, but it does not move over time, so it is not marked as possible object.}
	\label{fig_border_start-1}
\end{figure}

Algorithm~\ref{alg_preprocessing} describes the main preprocessing steps.
It aims at reasonable \initpoints{} to start object extraction and spatial borders ideally representing object silhouette bounds.
\begin{algorithm} %
	\caption{Preprocessing of the \EMAGS{}} %
	\label{alg_preprocessing} %
	\begin{algorithmic} %
		\renewcommand{\algorithmicrequire}{\textbf{Input:}}
		\renewcommand{\algorithmicensure}{\textbf{Output:}}
		\REQUIRE \EMAGS{}
		\ENSURE Border mask and list of \initpoints{}
		\STATE Spatial and temporal smoothing
		\FOR {each time step}
		\STATE Extract edges
		\STATE Generate possible object border lines
		\STATE Calculate center points
		\ENDFOR
		\RETURN border mask and \initpoints{}
	\end{algorithmic}
\end{algorithm}
The \EMAGS{} is first smoothed with a $3$D Gaussian in $P_\Occupied(\East, \North, t)$.
The first and second derivative is calculated along all $3$ dimensions to obtain points of inflections spatially and temporally. 
Similar to edge detection the found points represent sinks and raises of $P_\Occupied(\East, \North, t)$.
We consider the found points as border mask in spatial domain.
In time domain, for each cell time steps $P_\Occupied(t)$ within a raise and a slope, as illustrated by the plot in Fig.~\ref{fig_emags-1}, are considered as traversed by a moving object.
We use cluster centers of these points as \initpoints{} for the extraction algorithm explained in the following sections.
One slice, i.e. time step, of the preprocessing result is shown in Fig.~\ref{fig_border_start-1} including a zoomed view showing \initpoints{} in detail. %
The border mask is plotted in blue, where each marked point is part of the border of a possible object.
The green points are the \initpoints{} marking an inner point of a possible object.
It is possible for an object to have multiple or no \initpoints{} in a specific time step, as the preprocessing is a coarse first evaluation.
However, every object that has a clear appearance at least once in the sequence will be marked with an \initpoint{} in that time step.
This data is the output of preprocessing and will be used in the main algorithm to extract actual objects with their correct shapes.
At this point, there is no temporal connection established between the \initpoints{}, as it is not clear if every \initpoint{} marks an actual object.
\section{Algorithm Overview}
\label{sec_algo_overview}
\begin{figure*}[h]
	\vspace{1.5mm}
	\centering
	\includegraphics[width=\linewidth]{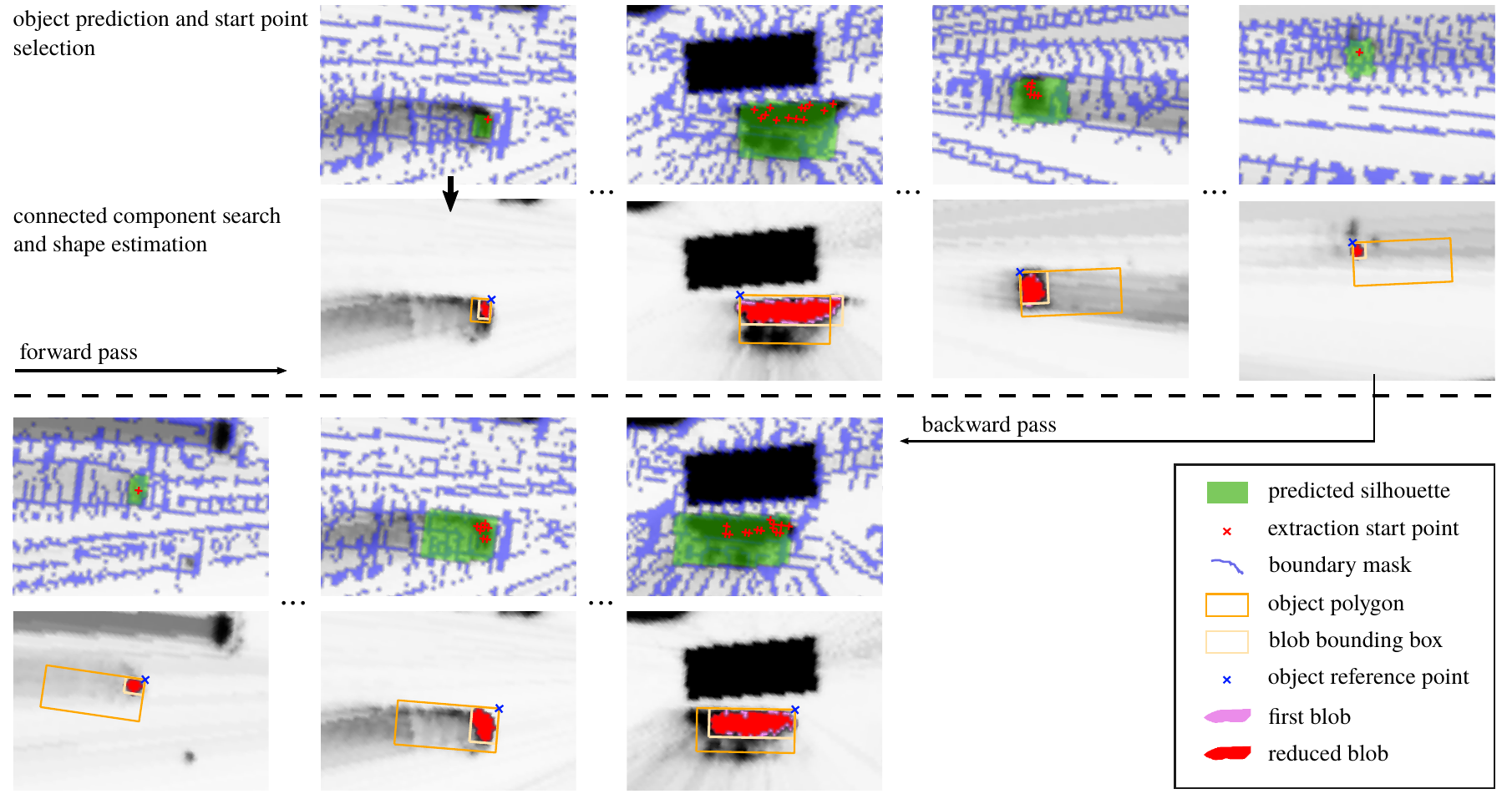}
	\caption{Overview over the introduced algorithm. First and third row show the prediction step including the predicted object silhouette as green rectangle and the calculated starting points for the next component search in red crosses. The second and fourth row show the result of the component search and the object pose estimation. The pink and red marked cells are the first and the reduced blob. The object's reference point is marked as blue cross. The beige and orange rectangles are the blob bounding box and the object bounding box, respectively.}
	\label{fig_algorthm_overview}
\end{figure*}
The present algorithm automatically generates object labels in the \EMAGS{} to enable their use as ground truth or comparison data.
Every object is traced through the considered sequence using the best fitting of length and width, which is obtained over multiple time steps.
The result is an automatically labeled \EMAGS{}, where ideally every occurring object has its correct dimension and position in every time step, even if the true dimensions are only observed in few time steps.
As the algorithm consists of multiple complex steps, this section gives an overview over the whole procedure, while the individual parts are explained separately in Sec.~\ref{sec_algo_components}.
\begin{algorithm} %
	\caption{Overview} %
	\label{alg_overview} %
	\begin{algorithmic} %
		\renewcommand{\algorithmicrequire}{\textbf{Input:}}
		\renewcommand{\algorithmicensure}{\textbf{Output:}}
		\REQUIRE \EMAGS
		\ENSURE labeled \EMAGS
		\STATE preprocess \EMAGS{} to calculate \initpoints{} and border mask
		\WHILE {get \initpoint{}}
		\STATE Object initialization: connected component, polygon, velocity profile
		\STATE Start temporal search
		\FOR {forward step, backward step}
		\WHILE {in sequence \AND object plausible}
		\STATE Object silhouette prediction
		\STATE Get connected component search starting points %
		\STATE Extract connected component: first blob
		\STATE Blob reduction via outlier removal
		\STATE Calculate velocity profile
		\STATE Object plausibility check
		\STATE Construct blob polygon and get reference point
		\STATE Update object width and length estimation
		\STATE Construct object polygon
		\ENDWHILE
		\STATE Start backward step with best object estimates from forward step
		\ENDFOR		
		\STATE Delete \initpoints{} covered by extracted object
		\STATE Object and trajectory consistency validation 
		\STATE Orientation correction for standing objects
		\STATE (optional) Temporal trajectory smoothing
		\STATE Write object to result
		\ENDWHILE
		\RETURN labeled \EMAGS
	\end{algorithmic}
\end{algorithm}
The pseudocode in Algorithm~\ref{alg_overview} introduces the idea of the main processing steps.
Fig.~\ref{fig_algorthm_overview} shows the main steps in detail in four rows of example pictures.
The first two rows illustrate the forward pass, while backward processing is depicted in the two bottom rows.
The first row shows in green the predicted visible silhouette of the last object extraction drawn over a grayscale DOGMa, where dark pixels refer to high $P_{\Occupied}$. 
A red cross illustrates cells within the predicted silhouette that fit best to the expected object velocity, $P_{\Occupied}$, and blob center.
These cells are used to start the connected component (blob) extraction.
Blue pixels refer to the current border mask limiting the connected component search.
The extracted connected component result is illustrated in the second row for each time step.
Please note, that first a rough blob (pink) is extracted based on previous object estimates, while a second, reduced blob (red) is obtained by outlier removal explained later in Section~\ref{subsec_algo_outlier_removal}.
A rectangle polygon is constructed around the reduced blob (light yellow rectangle).
The closest polygon point with least occlusion (sum of $P_{\Occupied}$ in line of sight) is considered as reference point (blue x).
The object size and length is estimated from current and previous blob polygons assuming up to \SI{10}{\percent} outlier probability.
The object polygon (orange rectangle) is constructed from the reference point and estimated object dimensions.
The third and fourth row show the same steps analogous, but in backward direction.
What should be noted is the already known object polygon in the backward phase that was calculated in the forward phase and would not be known from the measurement of the current blob.
Thus, correct object size and pose can be obtained even in far distance when the visible silhouette is corrupted due to particle convergence delay and (self-) occlusion.
\section{Algorithm Components}
The keywords used in Algorithm~\ref{alg_overview} are explained in this section.
Since fully detailed code would break the scope of the paper, all methods are also explained as pseudocode or described with few words.
\label{sec_algo_components}
\subsection{\InitPoint{}}
\label{subsec_algo_get_starting_point}
Each object initialization is based on a given \initpoint{} which is calculated by and obtained from the preprocessing.
As a result of the preprocessing, each \initpoint{} marks a moving object at some point in the sequence.
That means, an object does not need to have an \initpoint{} in each time step of the sequence, nor does it certainly have  only a single point.
As every \initpoint{} is as likely an object as another, all points generated in the preprocessing are put on a stack that is processed one by one.
However, as explained in Sec.~\ref{subsec_algo_remove_object_from_list}, all points covered by an object with completely examined trajectory are removed from the stack and do not spawn another new object.
\subsection{Velocity Profile}
\label{subsec_algo_vel_profile}
The \emph{velocity profile} of an object describes its characteristics statistically over cells occupied by the object.
As every cell holds information about its velocity, divided in east-/north-direction, each with the corresponding covariance, the resulting velocity vector can be calculated to provide an orientation and a velocity magnitude, as well as the corresponding covariance.
All valid cells included in one object, i.e. in one connected component, are used to retrieve the velocity profile.
Obviously invalid cells, i.e. cells that do not provide a valid covariance, are discarded to get as good a result as possible. 
The resulting velocity profile is used to distinguish incoming cells whether they fit in the object or not.
The velocity profile contains object wide features as well as cell wise features over all cells $c\in\mathcal{C}_0$, where $\mathcal{C}_0$ is the connected component occupied by the object.
The object wide features contain for $\EorN \in \left\{\East,\North \right\}$ the object mean velocity
\begin{equation*}
\bar{v}_{\EorN} = \bar{\sigma}_{v_\EorN} \sum_{c\in\mathcal{C}_0}\frac{1}{\sigma^2_{v_\EorN}(c)} v_{\EorN}(c) ~,
\end{equation*}
the object wide velocity variance
\begin{equation*}
\sigma_{\bar{v}_\EorN} = \left(\sum_{c\in\mathcal{C}_0}\frac{1}{\sigma^2_{v_\EorN}(c)}\right)^{-1} ~,
\end{equation*}
the object wide mean orientation
$
\OBJOrientation = \mathrm{atan2}( \bar{v}_{\North}, \bar{v}_{\East}) 
$
and velocity magnitude
$
\left|\bar{v}\right| = \sqrt{\bar{v}_{\North}^2 + \bar{v}_{\East}^2}~.
$
The cell wise statistics contain, over all object cells $c\in\mathcal{C}_0$, mean and variance of ${v}_{\East}(c)$, ${v}_{\North}(c)$, $\GMCellOrientation(c) = \mathrm{atan2({v}_{\North}(c), {v}_{\East}(c))}$, and $\left|{v}(c)\right| = \sqrt{{v}_{\North}(c)^2 + {v}_{\East}(c)^2}$.
The expected velocity variance in an object cell is calculated by
\begin{equation*}
\bar{\sigma}_{v_\EorN} = \frac{1}{\left|\mathcal{C}_0\right|}\sum_{c\in\mathcal{C}_0}\sigma_{v_{\EorN}}(c) ~.
\end{equation*}
Object wide features are used when assessing the object trajectory, while cell wise features are used find associating cells, e.g. in the next time step.
\subsection{Object Initialization}
\label{subsec_algo_init}
The \emph{object initialization}-method is used to calculate the first object state estimate based on the preprocessed data.
Algorithm~\ref{alg_init} describes the process of initializing a new object based on a given \initpoint{}.
The method is called for each \initpoint{} taken from the stack, while the \initpoint{} is required to have $\sigma^2_{v_\East}, \sigma^2_{v_\North} < 1\frac{\mathrm{m}^2}{\mathrm{s}^2}$ to ensure low uncertainty. 
The method uses a coarse-to-fine approach where the velocity profile and the connected component (see section \ref{subsec_algo_con_component}) are calculated twice in alternating order.
\begin{algorithm} %
	\caption{Object initialization} %
	\label{alg_init} %
	\begin{algorithmic} %
		\renewcommand{\algorithmicrequire}{\textbf{Input:}}
		\renewcommand{\algorithmicensure}{\textbf{Output:}}
		\REQUIRE \initpoint{} in space and time (E,N,t)
		\ENSURE Observed object (velocity profile, connected component, object polygon)
		\IF {Check \initpoint{} for prerequisites}
		\STATE Generate coarse connected component
		\STATE Generate coarse velocity profile
		\STATE Calculate fine connected component
		\STATE Calculate fine velocity profile
		\ENDIF
	\end{algorithmic}
\end{algorithm}
The result is an object hypothesis comprising connected grid cells, a velocity profile, and a bounding polygon.
From this hypothesis the object is traced forward and backward in time, as described in the following.
\subsection{Object Prediction}
\label{subsec_algo_obj_prediction}
The object prediction works in two ways, on object polygon level and on cell cluster (blob) level.
In early stages of the algorithm, both levels may be very similar, since the object size is similar to the connected component size, as no further information from other time steps is present.
In later stages, the knowledge of the object's dimension enables the tracing of a larger object than actual visible in the grid map as blob, e.g. due to (self) occlusion.
Therefore, the object polygon is predicted with constant velocity, with the prediction area increased by the variance in the velocity profile.
Thereby, the possible occupied cells of the whole object are found out.
Additionally, the visible blob is also predicted with constant velocity to obtain not only possible cells covered by an object but also cells expected to be visible as occupied.
This results in the possible positions of the actual measurable cells in the time step.
\subsection{Component Search Starting Point}
\label{subsec_algo_new_search_points}
After the prediction of an object and the resulting search space in the new time step, starting points for the connected component search are calculated.
The selection of those points aims at finding points fitting best to the expected blob size and velocity profile. 
The number of search start points is limited to one point per \SI{0.5}{m^2} of the object silhouette.
The selection is based on a loss function for every cell in the search space.
This loss function includes the following properties:
\begin{itemize}
	\item occupancy value
	\item orientation deviation from velocity profile
	\item distance deviation from expected blob center
	\item velocity deviation from velocity profile
\end{itemize}
The differences are calculated according to the properties from the earlier processing time step.
The points that minimize the loss function, i.e. fit best to the earlier estimation and the prediction of the object, will be the new centers from which the new connected component will be generated.
\subsection{Connected Component}
\label{subsec_algo_con_component}
\begin{algorithm}[h] %
	\caption{Connected component search} %
	\label{alg_con_comp_search} %
	\begin{algorithmic} %
		\renewcommand{\algorithmicrequire}{\textbf{Input:}}
		\renewcommand{\algorithmicensure}{\textbf{Output:}}
		\REQUIRE component search start points $s_0$, border mask $\mathcal{B}_0$, velocity profile $\mathcal{V}_0$
		\ENSURE connected component $\mathcal{C}_0$
		\STATE Stash $\mathcal{S} \gets s_0$
		\WHILE {$\mathcal{S} \ne \emptyset$}
		\STATE Pop $s_i$ from $\mathcal{S}$
		\STATE Add cells surrounding $s_i$ to $\mathcal{S}$ and to $\mathcal{C}_0$
		\STATE Remove already visited cells from $\mathcal{S}$
		\STATE Remove cells on $\mathcal{B}_0$ from $\mathcal{S}$
		\STATE Remove cells below occupancy threshold  from $\mathcal{S}$
		\STATE Remove cells not matching  $\mathcal{V}_0$  from $\mathcal{S}$
		\ENDWHILE
		\RETURN $\mathcal{C}_0$
	\end{algorithmic}
\end{algorithm}
In this context, a \emph{connected component} is a hypothesis which cells may belong to an object.
Starting from an \initpoint{} or component search start point it grows successively by adding adjacent cells until it reaches a boundary provided by the border mask.
Whereas, cells that
\begin{itemize}
	\item lie on the border,
	\item fall below an occupancy threshold,
	\item do not match the velocity profile
\end{itemize}
are discarded from further exploration. 
The thresholds are controlled by the predicted silhouette and velocity profile, e.g. a $\pm2\sigma$ band from the velocity profile around mean orientation of component search start points, and a $-2\sigma$ band from mean $P_\Occupied$ accordingly. 
Algorithm~\ref{alg_con_comp_search} describes the connected component search regarding the border mask and the velocity profile.
Note that all surrounding points of a stashed point are added to the connected component $\mathcal{C}_0$ but only the points meeting the required properties are added as additional search points to the stash $\mathcal{S}_0$.
Therefore, the resulting connected component consists of inner points matching the velocity profile and a maximum of one layer of boundary points that may violate the velocity profile. 
\subsection{Outlier Removal}
\label{subsec_algo_outlier_removal}
\begin{figure}[h!]
	\vspace{2mm}
	\centering
	\includegraphics[width=0.9\linewidth]{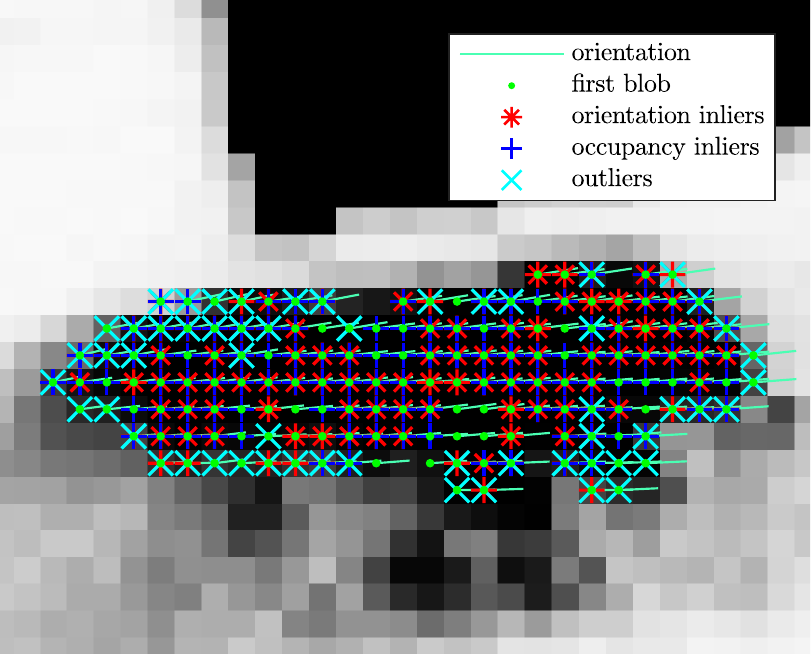}
	
	\vspace{4mm}
	
	\includegraphics[width=0.9\linewidth]{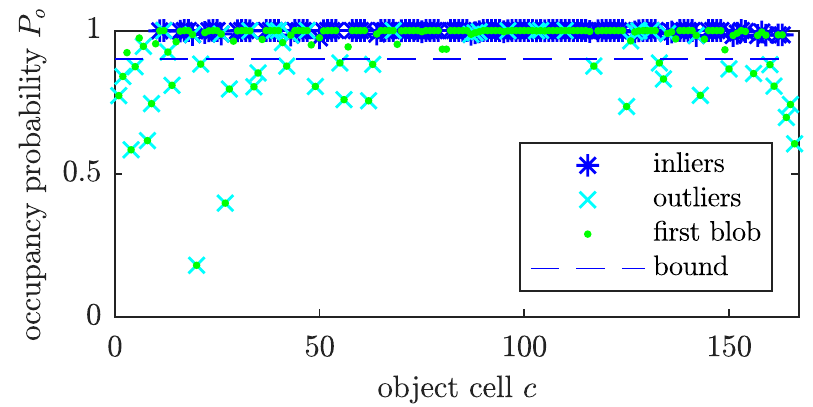}
	
	\vspace{3mm}
	
	\includegraphics[width=0.9\linewidth]{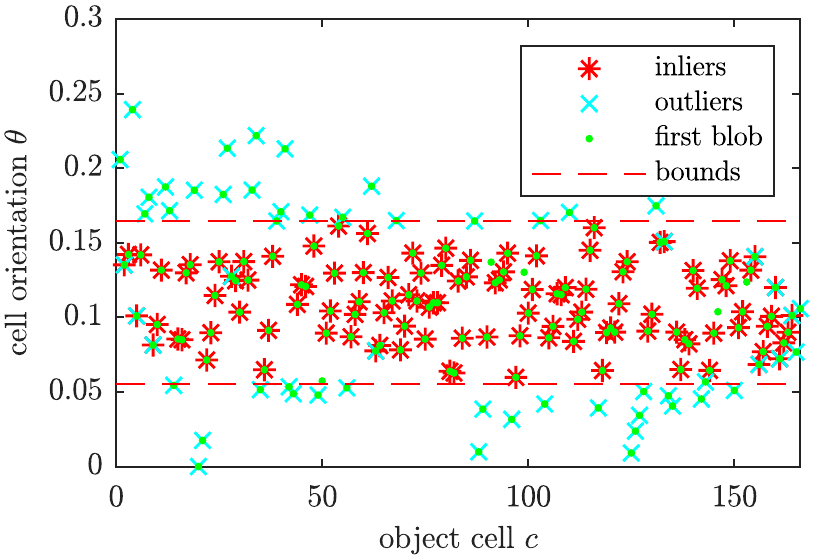}
	\caption{Blob shrinking by outlier removal.
		The first extracted blob is assumed to contain $n$ inliers, while $n$ is the number of previous blob cells.
		$n$ cells with highest $P_{\Occupied}$ and minimal deviation from the mean orientation are considered as inliers and used to calculate outlier bounds. 
		Cells not identified as outliers are included in the reduced blob.
		This includes also cells not considered as inlier nor outliers.
	}
	\label{fig_outlierspage}
\end{figure}
The calculated connected component, based on the starting points from the prediction step, is assumed to include outliers, as the connected component search aims on finding all possible object cells suiting the previous object state.
This first connected component is called \emph{first blob} in Fig.~\ref{fig_algorthm_overview} and outlier removal leads to the \emph{reduced blob}, shown in the same figure.
In the removal step only the cells certainly belong together should be taken into account for the shape estimation.
The considered properties are
\begin{itemize}
	\item occupancy value
	\item orientation value
	\item velocity value
\end{itemize}
of every cell in the first blob which results in a mean value and a standard deviation for each property.
All cells that lie out of a two-sigma band, i.e. differ more than two standard deviations from the mean, are removed as outliers from the blob.
Fig.~\ref{fig_outlierspage} illustrates the outlier removal in one silhouette.
From the previous time step it is known, that the blob consists of $n$ cells. 
Therefore it is assumed, that the new blob contains $n$ inliers. 
From the new blob, $n$ cells with highest $P_\Occupied$ and lowest deviation from mean orientation $\GMCellOrientation$ are used to calculate the standard deviation, which in turn is used to separate outliers from the first blob.
The resulting blob might than contain more than $n$ cells.
Please note, that the outlier bounds are limited to a minimum band, i.e. expected variance in the velocity profile and $0.9\cdot\underset{c\in\mathcal{C}_0}{\mathrm{max}}\left(P_\Occupied(c)\right)$.
\subsection{Removal of completed object}
\label{subsec_algo_remove_object_from_list}
A fully examined and saved object has to be removed from the searching list.
As one object may cover multiple \initpoints{} in each time step of the \EMAGS{}, every affiliated point needs to be removed, spatial as temporal.
Algorithm~\ref{alg_obj_rem} explains how completed objects are removed from the list of \initpoints{}.
\begin{algorithm} %
	\caption{Object Removal} %
	\label{alg_obj_rem} %
	\begin{algorithmic} %
		\renewcommand{\algorithmicrequire}{\textbf{Input:}}
		\renewcommand{\algorithmicensure}{\textbf{Output:}}
		\REQUIRE Object, preprocessed \EMAGS{} %
		\ENSURE \EMAGS{} without input object
		\STATE Extract object dimension
		\STATE Transform object in every relevant time step
		\STATE Determine underlying cells
		\STATE Remove cells from possible \initpoints{}
	\end{algorithmic}
\end{algorithm}
This step ensures that the algorithm terminates, as it removes at least the \initpoint{} that was considered as possible object.
Additionally, as an object is generated by a single \initpoint{}, but may overlap multiple \initpoints{} over different time steps, this step commonly removes more than one point from the stack.
Thereby, the calculation time, dependent on the amount of initialized objects, is reduced heavily.

\subsection{Post Processing}
\label{subsec_algo_post_processing}
The extracted object trajectory is evaluated for plausible size, shape aspect ratio and smooth movement.
Also, trajectories traversing buildings permanently are ignored, while short inference with buildings is tolerated due to localization and map uncertainties.
Objects within buildings are usually caused by mirrored laser measurements at glass fronts of buildings.
Buildings are represented as polygons obtained from Open Street Maps.
It happens that the algorithm traces standing objects.
Therefore, static trajectories are ignored.
In addition, orientation estimation of objects temporarily standing is error prone and thus corrected using linear interpolation where the trajectory doesn't move.
\section{Results}
\label{sec_results}
\begin{figure}[t]
	\vspace{2mm}
	\centering
	\includegraphics[width=0.8\linewidth]{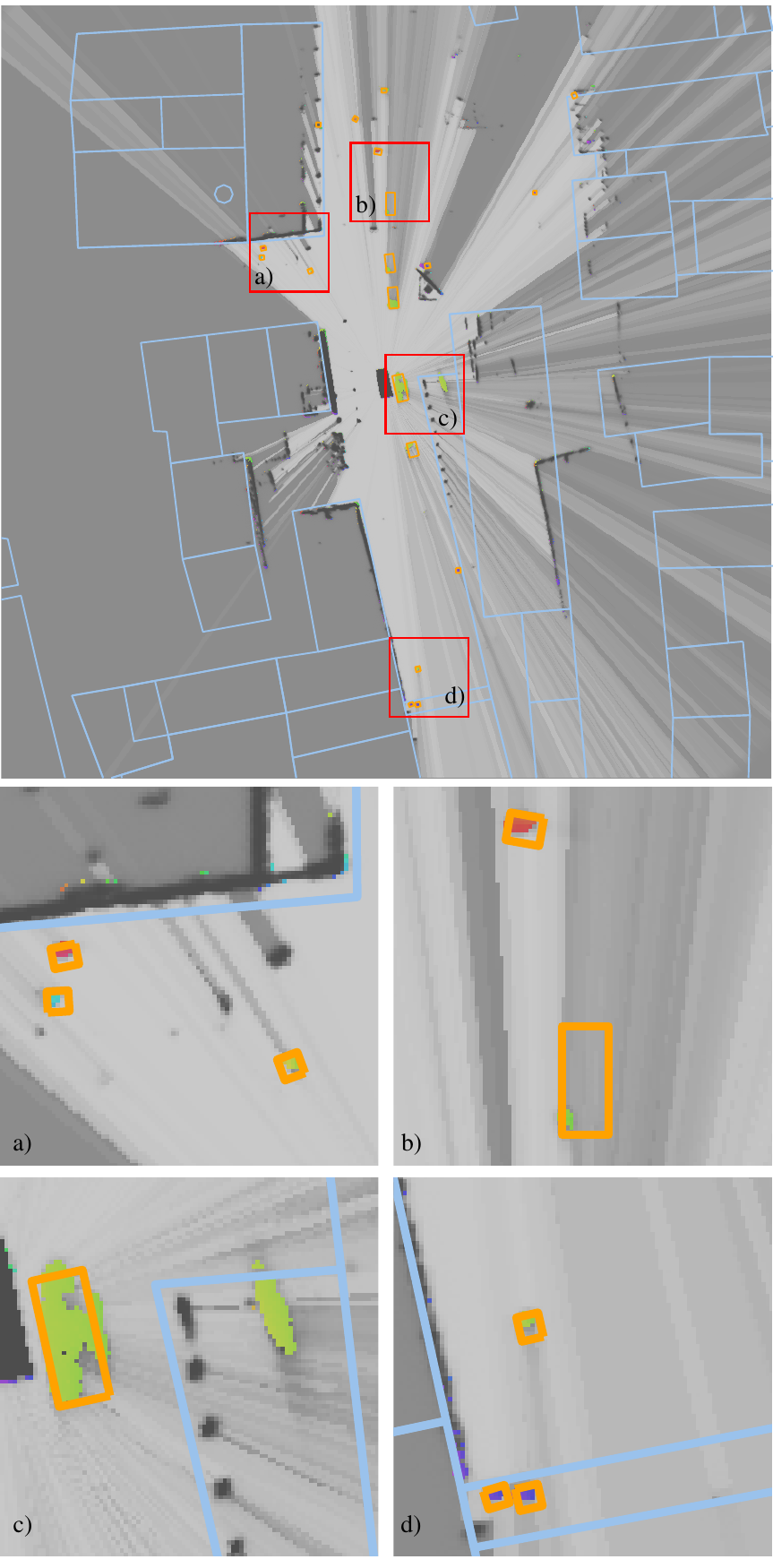}
	\caption{Example scene of extracted objects with zoomed excerpts a)-d). The occupancy values are illustrated in the background, where moving cells are colored according to their direction. The extracted object labels are shown with orange rectangles. Buildings are plotted with light blue lines, taken from the open street map. 
	}
	\label{fig_resultsexample}
\end{figure}
\begin{figure}[t]
	\vspace{2mm}
	\centering
	\includegraphics[width=0.9\linewidth]{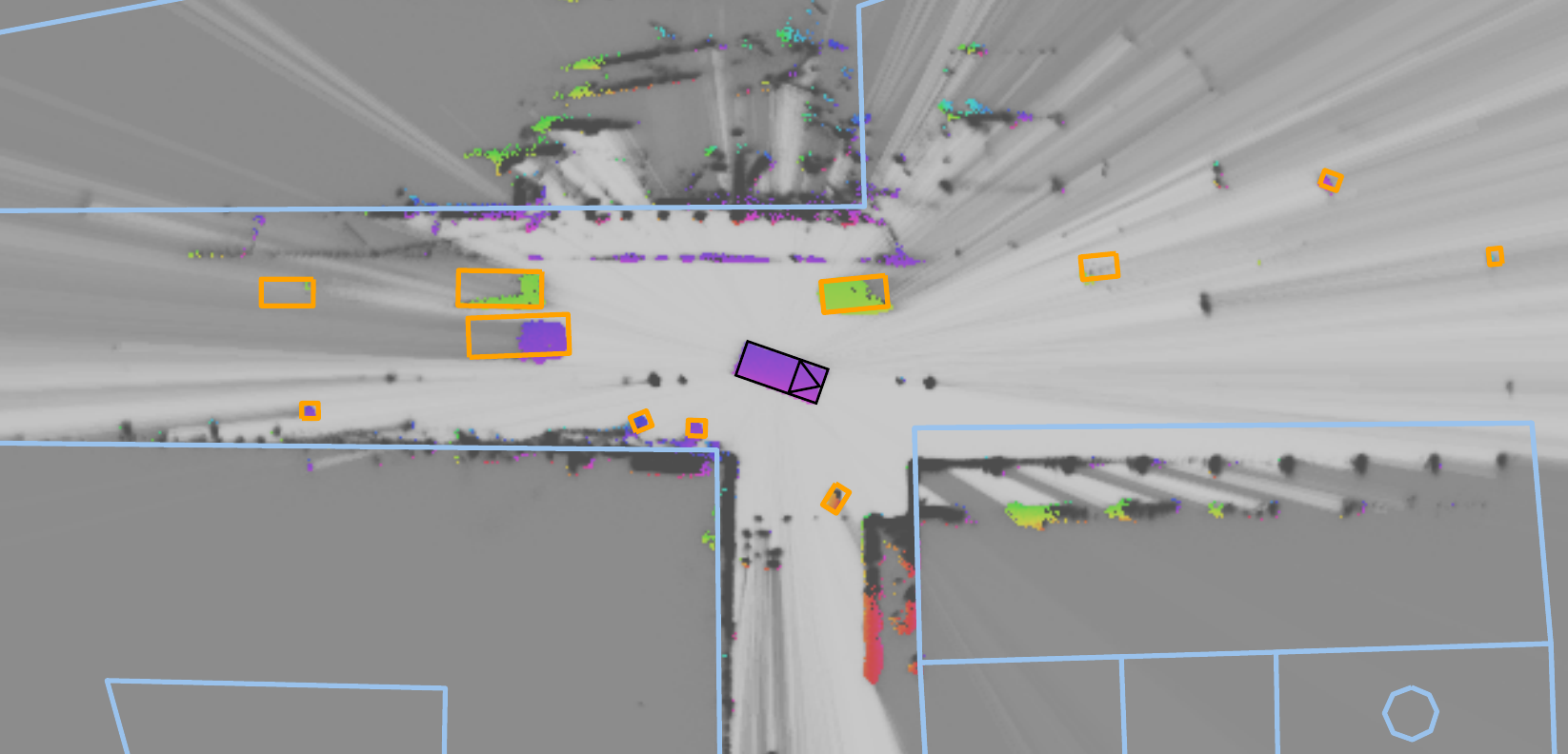}
	\caption{Example scene of extracted objects while the ego vehicle (black rectangle with inner triangle) is moving. 
	}
	\label{fig_resultsexample_moving}
\end{figure}
The algorithm was applied on laser recordings from a down town shared space scenarios including multiple pedestrians, bikes, cars, trucks and buses. 
The experimental vehicle is equipped with multiple laser scanners, four 16-layer Velodyne scanners and one 4-layer Ibeo Lux. 
The scene was recorded for about \SI{2.5}{h}.
Although recordings were made with a moving and stationary platform, due to the high traffic, most of the sequence was recorded from a parking position either in the street center or on the sidewalk.
As the presented method generates labels thought as ground truth data, it has to compete with manual labeling and thereby is best validated visually.
Fig.~\ref{fig_resultsexample} shows some examples where the generated object rectangles are plotted in orange, open street map buildings in blue, moving cells in colors according to their direction and the occupancy values in shades of gray. 
The zoomed excerpts are: a) Three objects (pedestrians) are extracted correctly. b) Two objects (pedestrian and vehicle) are extracted, where the current grid map state would not lead to the correct vehicle size. c) State estimation of the left vehicle fits to the measured cells. The mirrored blob in the right building is omitted, because its trajectory lies inside the building. d) Three pedestrians are correctly extracted, although they are far away from the ego vehicle and close together, which would typically result in one large detection or no detection at all.
An example where the ego vehicle is moving is illustrated in Fig.~\ref{fig_resultsexample_moving}.
The example shows, that many static regions in the grid map have a false velocity estimation, illustrated by colored grid map pixels.
The EMAGS offline assessment, however, resolves that the occupancy is actually not moving although the particle filter indicates dynamic states.

The main limitation of the algorithm is that if track of an object is lost due to temporary full occlusion, reinitialized object tracing easily fails to estimate the correct object size.
\section{Conclusion}
\label{sec_conclusion}
A new method to generate object labels on a DOGMa is introduced in this work.
After the preprocessing of a DOGMa sequence, called \EMAGS, the algorithm uses a forward search to find objects and calculate their dimensions and poses.
Additionally a backward phase is used to predict the object back to the beginning of the sequence, whereby the best possible object estimation is achieved.
The extracted object dimensions and poses serve as automatically generated ground truth labels in the DOGMa.
The advantage of this method is that the labels are generated automatically and not manually, thereby it is possible to label almost every amount of sequences, only limited through computation time and not through persons labeling the single images. 

\bibliographystyle{IEEEtran}
\balance
\bibliography{IEEEabrv,bib/myBibTex}

\end{document}